\PassOptionsToPackage{table}{xcolor}
\documentclass[letterpaper]{article} 
\usepackage{aaai2026}  
\usepackage{times}  
\usepackage{helvet}  
\usepackage{courier}  
\usepackage[hyphens]{url}  
\usepackage{graphicx} 
\usepackage{natbib}  
\usepackage{caption} 
\usepackage{amsmath}
\usepackage[utf8]{inputenc} 
\usepackage{booktabs}       
\usepackage{amsfonts}       
\usepackage{nicefrac}       
\usepackage{microtype}      
\usepackage{lipsum}		
\usepackage{graphicx}
\usepackage{natbib}
\usepackage{multirow}
\usepackage{tikz}
\usetikzlibrary{arrows.meta, positioning}
\usepackage{subcaption}
\usepackage{bm}
\usepackage{textcomp}

\usepackage[ruled,linesnumbered,vlined]{algorithm2e}
\SetAlCapHSkip{0ex}
\usepackage{etoolbox}
\makeatletter
\patchcmd\algocf@Vline{\vrule}{\vrule \kern-0.4pt}{}{}
\patchcmd\algocf@Vsline{\vrule}{\vrule \kern-0.4pt}{}{}
\makeatother

\SetNlSty{large}{}{:}
\usepackage{listings}
\usepackage{xcolor}

\definecolor{eclipseStrings}{RGB}{42,0.0,255}
\definecolor{eclipseKeywords}{RGB}{127,0,85}
\colorlet{numb}{magenta!60!black}
\definecolor{lycolor}{RGB}{0,128,128}      

\definecolor{rzycolor}{RGB}{150,75,0}

\usepackage{booktabs,arydshln}
\makeatletter
\def\adl@drawiv#1#2#3{%
        \hskip.5\tabcolsep
        \xleaders#3{#2.5\@tempdimb #1{1}#2.5\@tempdimb}%
                #2\z@ plus1fil minus1fil\relax
        \hskip.5\tabcolsep}
\newcommand{\cdashlinelr}[1]{%
  \noalign{\vskip\aboverulesep
           \global\let\@dashdrawstore\adl@draw
           \global\let\adl@draw\adl@drawiv}
  \cdashline{#1}
  \noalign{\global\let\adl@draw\@dashdrawstore
           \vskip\belowrulesep}}
\makeatother
\usepackage{amssymb}
\usepackage{pifont}
\usepackage{marvosym}

\title{Faithful in Steps: Improving Generalization and Citation in RAG \\via Query Decomposition}

\author {
    Yue Liu\textsuperscript{\rm 1}\footnotemark[1],
    Zhongying Ru\textsuperscript{\rm 2},
    Shimin Di\textsuperscript{\rm 3,\rm 4},
    Jipeng Zhang\textsuperscript{\rm 1},
    Ruiyuan Zhang\textsuperscript{\rm 5},
    Xiaofang Zhou\textsuperscript{\rm 1}\thanks{Corresponding authors.}
}
\affiliations {
    \textsuperscript{\rm 1}The Hong Kong University of Science and Technology (HKUST)\\
    \textsuperscript{\rm 2}Independent Researcher \\
    \textsuperscript{\rm 3}School of Computer Science and Engineering, Southeast University\\
    \textsuperscript{\rm 4}Key Laboratory of New Generation Artificial Intelligence Technology and Its Interdisciplinary Applications\\ 
    \textsuperscript{\rm 5}Hong Kong Generative AI Research and Development Center (HKGAI)\\
    yliumh@connect.ust.hk, zxf@ust.hk
}

\begin{document}

\maketitle

\begin{abstract}
Retrieval-augment generation is a prevalent strategy to mitigate hallucinations of LLMs. The attributable RAG (RAGQ) generates quotes for its answers. The quotes indicate which input contexts support the RAG to derive the answers, enhancing the answer's verifiability and trustworthiness.
However, existing RAGQs exhibit significant degradation when dealing with questions that require multi-hop reasoning and multi-modal understanding, suffering from over-citation, implicit entity identification failure, and poor generalization.
In this paper, we propose a novel RAGQ framework, namely \textbf{QDRAG}. 
QDRAG breaks down the input question into atomic subquestions to identify the implicit entities. Then, the reranker prunes context distractors to eliminate the downstream over-citation. To facilitate query decomposition, we propose two zero-shot approaches: QD-C and QD-R, which guide the QD MLLM to decompose the question based on context knowledge and retrieval rewards, respectively.
One interesting finding is that finetuning on the QD task shows better generalizability compared to directly finetuning on the downstream RAGQ task. 
Experiments on four multi-modal QA benchmarks demonstrate QDRAG's efficacy in grounding answers and generating faithful citations. The framework significantly outperforms all the baselines on both in-domain and out-of-domain tests, even surpassing Gemini-Pro.
\end{abstract}


\section{Introduction}

Retrieval-augmented generation (RAG)~\cite{abootorabi2025ask} is a common method to mitigate hallucinations in large-language models.
RAG retrieves external data and grounds its answer on the retrieved contexts. 
Recent studies~\cite{huang-etal-2024-learning,song2025measuring} propose RAG with quotes (RAGQ) to improve answer faithfulness. 
RAGQ outputs include answers and verifiable source references that indicate which retrieved data support the LLM's outputs. Users can verify answers via cited sources, and this alignment between responses and sources is key to answer integrity.

\begin{figure}[t]
    \centering
    \includegraphics[width=\linewidth]{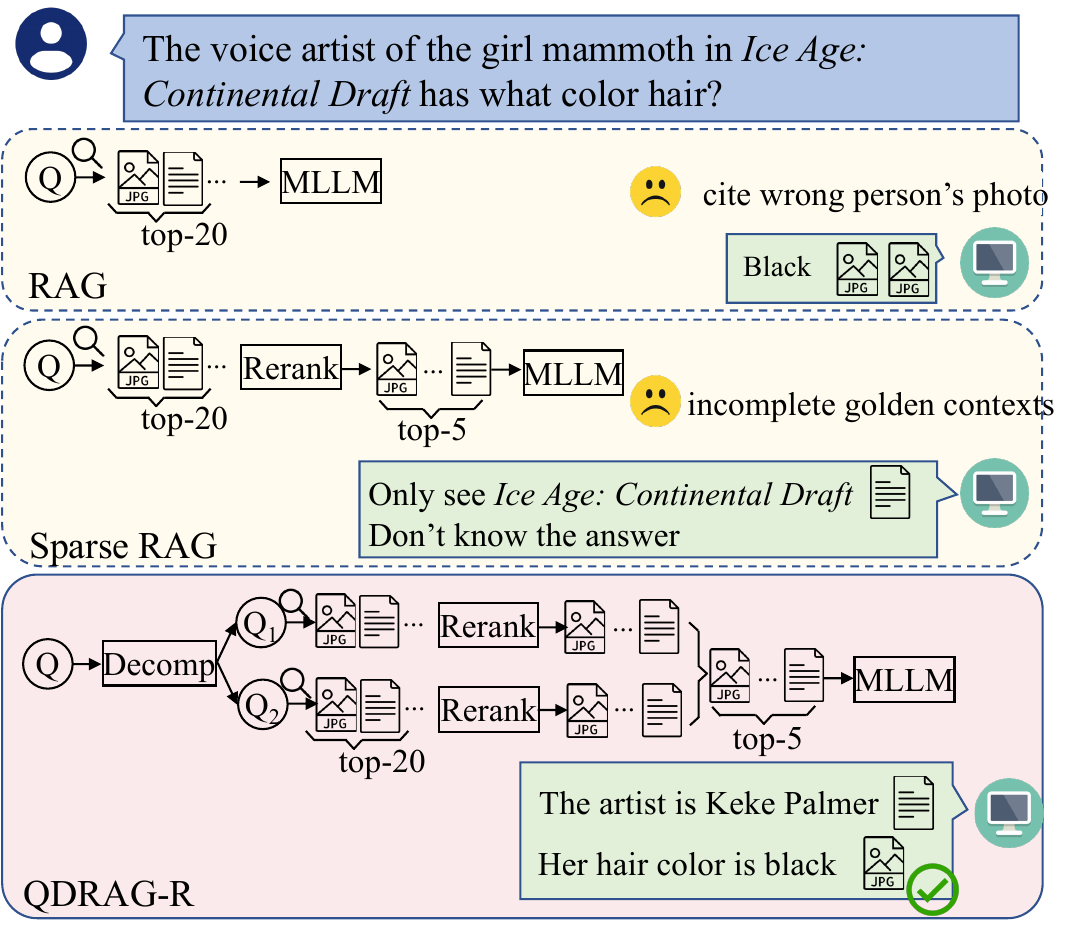}
    \caption{The massive retrieved data in RAG misleads the MLLM to cite the incorrect but similar contexts; Sparse RAG drops key contexts in the reranking step for multi-hop questions; QDRAG filters the distractors while maintaining the key contexts for accurate downstream citations.}
    \label{fig:intro}
    \vspace*{-2mm}
\end{figure}

Despite recent advancements, current RAGQs still face challenges for multi-hop and multi-modal questions.

\noindent \textbf{$\bm{\mathcal{C}}$-1: Over-citation of incorrect contexts}. The RAGQ cites multi-modal contexts where the local segments align with the answer but neglects to evaluate the broader relevance of these contexts. 
\noindent \textbf{$\bm{\mathcal{C}}$-2: Implicit entity retrieval failure}. The post-hoc retrieval approaches~\cite{gao-etal-2023-rarr} rely on retrievers to search for supporting contexts, but simply using the original text of the question as a query often fails due to the key entities being absent in the multi-hop question.
\noindent \textbf{$\bm{\mathcal{C}}$-3: Poor generalization ability.}
Instruction fine-tuning methods~\cite{huang-etal-2024-learning,song2025measuring} finetune LLMs for downstream tasks. They build training data to enhance LLMs' attribution capability, leading to difficulties in maintaining reference accuracy for dynamic multimedia knowledge~\cite{bi2024context}.

One intuitive solution is to use sparse RAG~\cite{zhu2025accelerating} to pre-filter contexts to a shorter list before MLLM generation, which assesses context relevance in parallel.
However, for multi-hop questions, sparse RAG misjudges the contexts with partial knowledge as ``irrelevant'' due to \textbf{$\bm{\mathcal{C}}$-2}. Another research line decomposes the question into a reasoning plan by few-shot in-context learning. The domain and difficulty gaps between examples and real questions make the MLLM hard to learn how to effectively simplify the multi-hop question~\cite{zhou2023leasttomost,khot2023decomposed}. 

To address the above challenges,
we propose a novel RAGQ framework named QDRAG, which consists of three main modules: query decomposer (QD), context reranker, and generator.
Different from the static contexts used in existing RAGQs, QDRAG filters distractors before answer generation. Sparse contexts help MLLMs focus on correct contexts and generate better quotes (\textbf{$\bm{\mathcal{C}}$-1}).
The QD-reranking paradigm assesses context relevance to single-hop subquestions, thereby resolving \textbf{$\bm{\mathcal{C}}$-2}.
For query decomposition, we propose zero-shot QD-C and QD-R, two new QD models that eliminate manually written in-context examples.  
QD-C breaks the QD task into two clearly defined operators: entity replacement and open decomposition. QD-C is plug-and-play for existing RAGs. 
QD-R uses the recall rate of golden contexts as the reward of a decomposition plan to perform direct preference optimization on the QD task. 
The retriever feedback significantly boosts the automatic annotation process compared to LLM feedback used by RAGQs. Moreover, the inputs of the decomposition task (a multi-hop question) are context-agnostic and answer-agnostic. The finetuned QDRAG thus does not suffer from the massive and dynamic multimedia contexts. The existence of common decomposition logic among multi-hop questions makes QDRAG more generalizable than existing RAGQs ($\bm{\mathcal{C}}$-3).
Our contributions are summarized as: 
\begin{itemize}
    \item We highlight the close relationship between context sparsity and MLLM context reference capability. It motivates our new QD-Reranking-based RAGQ pipeline.
    \item We propose two query decomposition methods to identify implicit entities on multi-hop questions, and subquestion-based context reranking to address the over-citation problem on multi-modal contexts. We enhance the model generalizability by finetuning QDRAG on the QD task.
    \item Extensive experiments demonstrate that QDRAG reduces answer and citation hallucinations and demonstrates remarkable generalization across different base models.
\end{itemize}


\section{Related Work}

\subsubsection{RAG with Quotes (RAGQ).}

The training-free RAGQs use post-hoc retrieval~\cite{gao-etal-2023-enabling}, chain-of-thought prompting~\cite{ji2024chain} or dynamic programming~\cite{choi2025trustworthy} to select supporting documents.
For training-based methods, \cite{asai2024selfrag,xia-etal-2025-ground} perform supervised-finetuning on RAG LLMs to enable context citations. 
\cite{huang-etal-2024-learning,song2025measuring} employs reinforcement learning with human feedback or direct preference optimization. 
Self-Cite~\cite{chuang2025selfcite} utilizes the change of LLM's confidence on an answer when removing the cited documents to calculate the context's importance.
In contrast to previous work, QDRAG addresses the RAGQ task by performing query decomposition and context filtering to reduce the citation candidates and hallucinations. 

\subsubsection{LLM for Reranking.}

There are four categories in this research line: pointwise, pairwise, listwise, and setwise. The pointwise approach evaluates context relevance one by one. The score can either be a binary label ``yes/no'', or the normalized likelihood of generating a ``yes'' response~\cite{zhu2025accelerating, yu2024rankrag}. Listwise methods prompt an LLM to output the ordered sequence~\cite{sun-etal-2023-chatgpt}. Pairwise prompts let the LLM decide which of two contexts is more relevant~\cite{qin-etal-2024-large}. 
Setwise approach directly outputs the relevant set of contexts~\cite{10.1145/3626772.3657813,yu2024rankrag}. Among the four categories, the pointwise method is the only one that is parallelizable and also has comparable performance with others~\cite{10.1145/3626772.3657813}.

\subsubsection{LLM for Query Decomposition.}

Least-to-Most~\cite{DBLP:conf/iclr/ZhouSHWS0SCBLC23}, Decomp~\cite{khot2023decomposed}, Dynamic Least-to-Most~\cite{DBLP:conf/iclr/DrozdovSASSCBZ23}, and Self-Ask~\cite{press-etal-2023-measuring} use in-context demonstrations to iteratively decompose the question and answer the subquestions step by step. 
Recently, SearChain~\cite{xu2024search} adopts the decomposition plan as the generation unit and iteratively refines the plan, surpassing PS~\cite{DBLP:conf/acl/WangXLHLLL23} on multi-hop question answering. 

\section{Methodology}

\begin{figure}[t]
    \centering
    \begin{subfigure}{\linewidth}
        \centering
        \includegraphics[width=\linewidth]{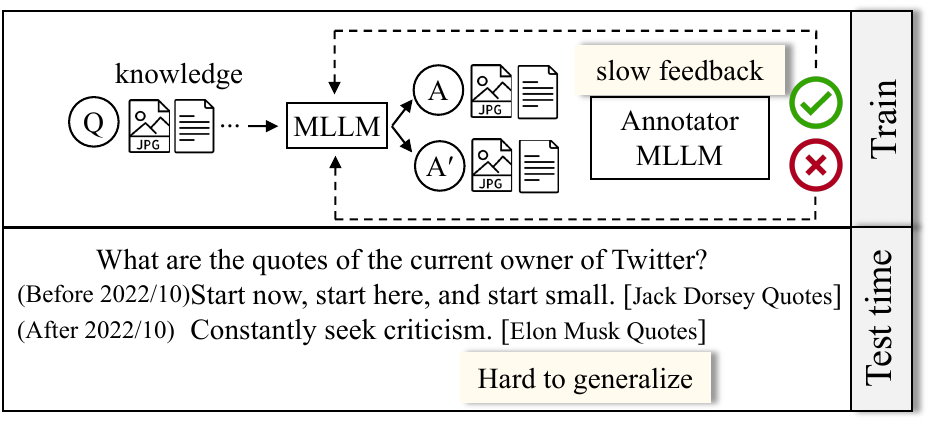}
        \caption{Finetuning on the RAGQ task}
        \label{fig:ft_ragq}
    \end{subfigure}
    \quad
    \begin{subfigure}{\linewidth}
        \centering
        \includegraphics[width=\linewidth]{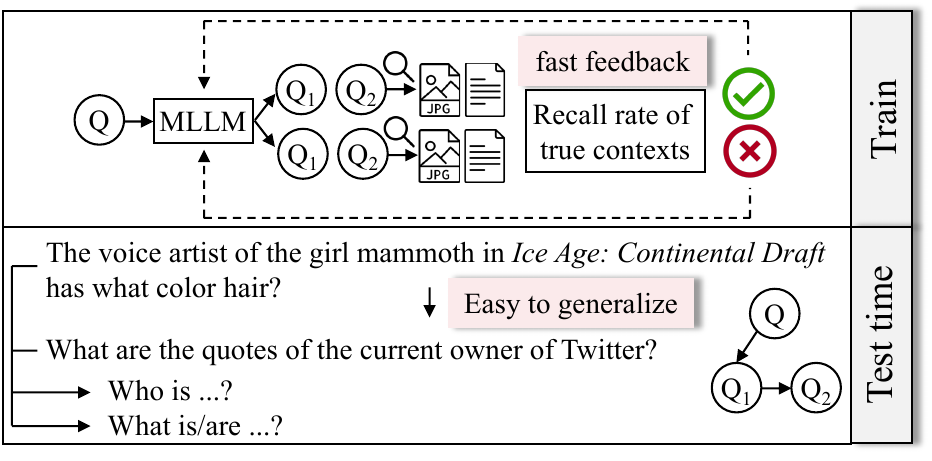}
        \caption{Finetuning on the QD task}
        \label{fig:ft_qd}
    \end{subfigure}
    \caption{QDRAG finetunes the MLLM for the QD task, leveraging its fast training and better generalizability.}
    \vspace*{-2mm}
\end{figure}

\subsection{Preliminaries}
\subsubsection{Vanilla Workflow of RAGQ.}

Given a user query $Q$, the multi-modal RAGQ retrieves visual and text knowledge pieces from external databases.
Then, it merges the knowledge pieces with $Q$ as the input context.
Next, a multi-modal large language model (MLLM) is prompted to produce an answer with in-line citations in the form of [1][2]. The goal of RAGQ (RAG with quotes) is to ensure that the answer is supported by the cited knowledge sources.

\subsubsection{Multi-hop Question Answering.}  
Answering a multi-hop question needs several reasoning steps~\cite{trivedi2022musique}. The question can be decomposed repeatedly until it can be answered by a single external document. 
A subquestion $Q_i$ can depend on the answer to another subquestion $Q_j$.
In Figure~\ref{fig:intro}, if $Q_i=\text{``Who is the voice artist?''}$, and $Q_j=\text{``}\langle\text{Answer\_of\_}Q_i\rangle \text{ has what color hair?''}$, then $Q_j$ relies on $Q_i$. 
There are also independent subquestions. For instance, the question ``Which film was released first: $\langle a\rangle$ or $\langle b\rangle$?'' has two independent subquestions: $Q_i=\text{``When was $\langle a\rangle$ released?''}$ and $Q_j=\text{``When was $\langle b\rangle$ released?''}$. 

\subsection{Sparse Context Benefits RAG Quotes} \label{sec:motivation}

We explore the impact of context sparsity on MLLM's reference performance through two tasks. 
Our findings show that the sparse RAGs generate better citations than dense RAGs.

\begin{figure}[t]
    \centering
    \begin{subfigure}{0.46\linewidth}
        \centering
        \includegraphics[width=\linewidth]{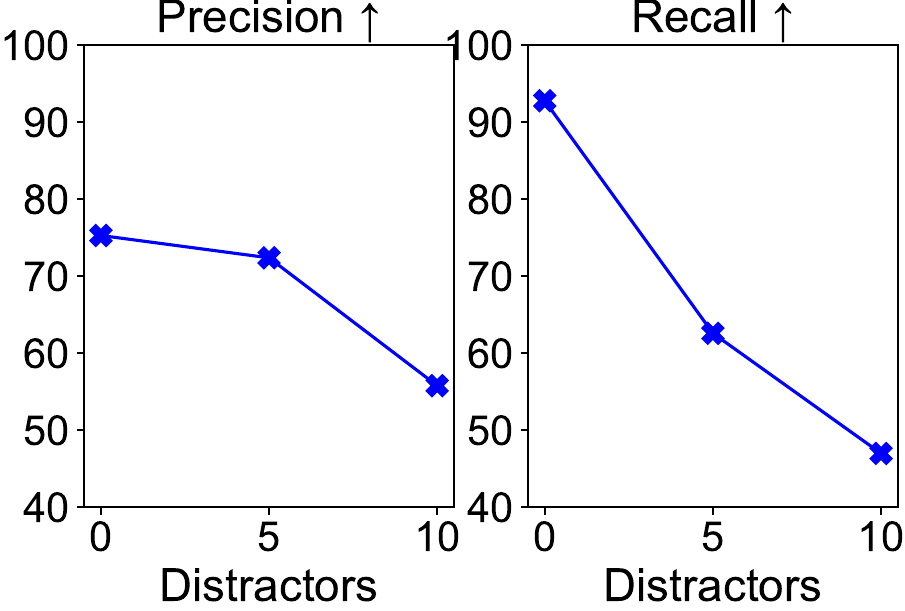}
        \caption{CIRCO}
        \label{fig:circo}
    \end{subfigure}
    \quad
    \begin{subfigure}{0.46\linewidth}
        \centering
        \includegraphics[width=\linewidth]{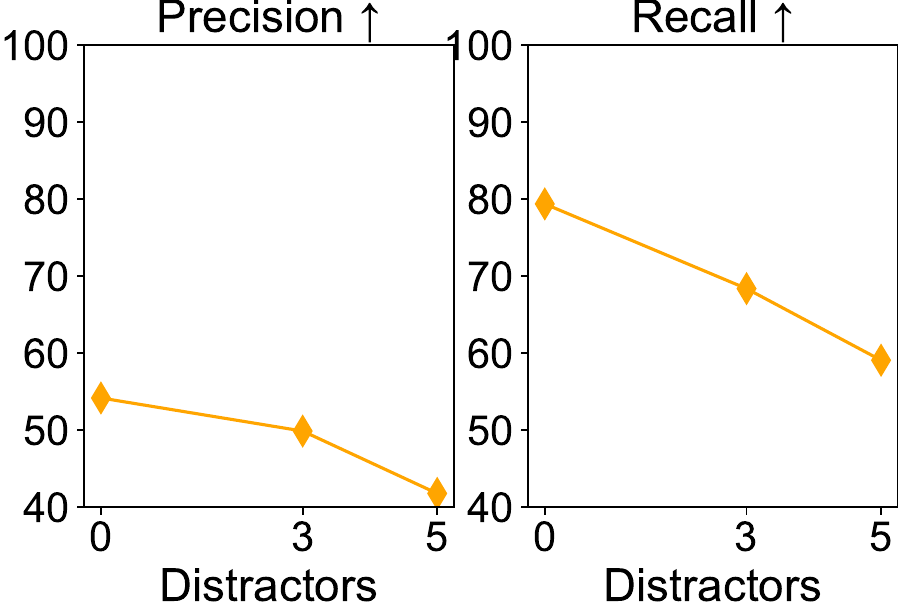}
        \caption{CIRR}
        \label{fig:cirr}
    \end{subfigure}
    \caption{Evaluate MLLM's capability to cite images from a multi-modal statement on two datasets: CIRCO and CIRR.}
    \label{fig:cir}
\end{figure}

The first task is to cite images from an expression of interleaved images and texts. It simulates the scenario that a RAGQ is about to generate citations after the answers. Specifically, the input template is ``Compared with the image $I_R$, the image has $\langle a \rangle$'' and a list of images ``[1] Image; [2] Image ...''. The outputs are image IDs (e.g., [1][3]) that support the statement. We test on two image retrieval datasets~\cite{baldrati2023zero,liu2021image}. In Figure~\ref{fig:cir}, the results show that reducing input distractors helps RAGQ reduce the risk of citing wrong images (higher precision) and cite a more complete context set (higher recall). Therefore, sparse RAG is a promising strategy to enhance RAG citations.

\begin{figure}[t]
    \centering
    \begin{subfigure}{0.45\linewidth}
        \centering
        \includegraphics[width=\linewidth]{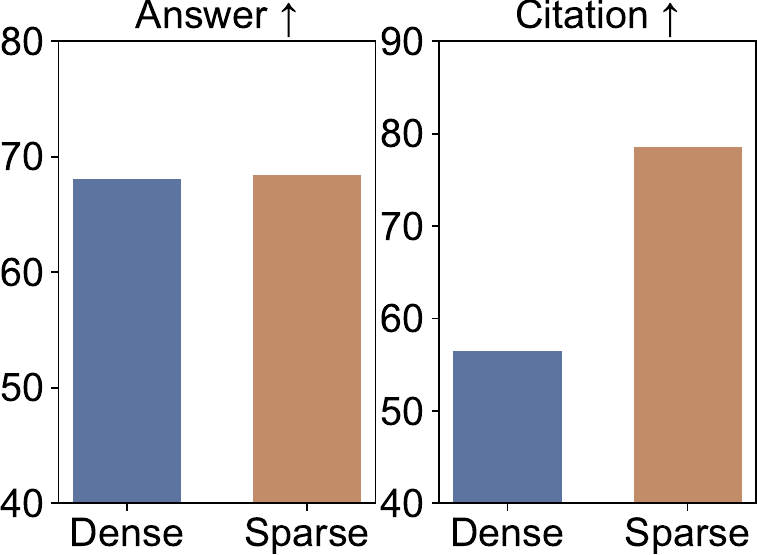}
        \caption{Single hop MMQA}
        \label{fig:single-mmqa}
    \end{subfigure}
    \quad
    \begin{subfigure}{0.45\linewidth}
        \centering
        \includegraphics[width=\linewidth]{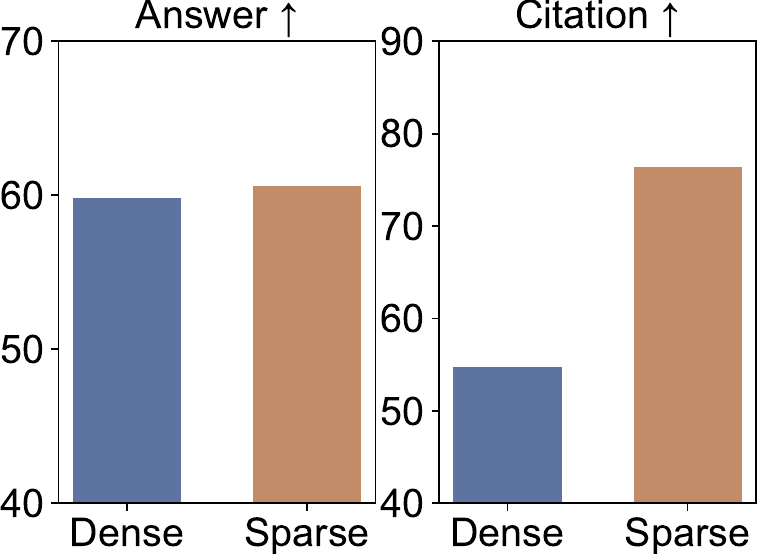}
        \caption{Single-hop WebQA}
        \label{fig:single-webqa}
    \end{subfigure}
    \\
    \begin{subfigure}{0.45\linewidth}
        \centering
        \includegraphics[width=\linewidth]{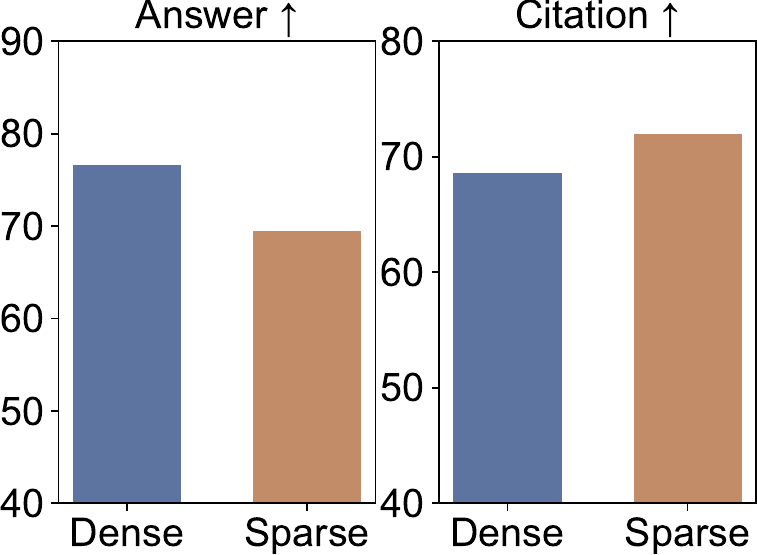}
        \caption{Multi-hop MMQA}
        \label{fig:multi-mmqa}
    \end{subfigure}
    \quad
    \begin{subfigure}{0.45\linewidth}
        \centering
        \includegraphics[width=\linewidth]{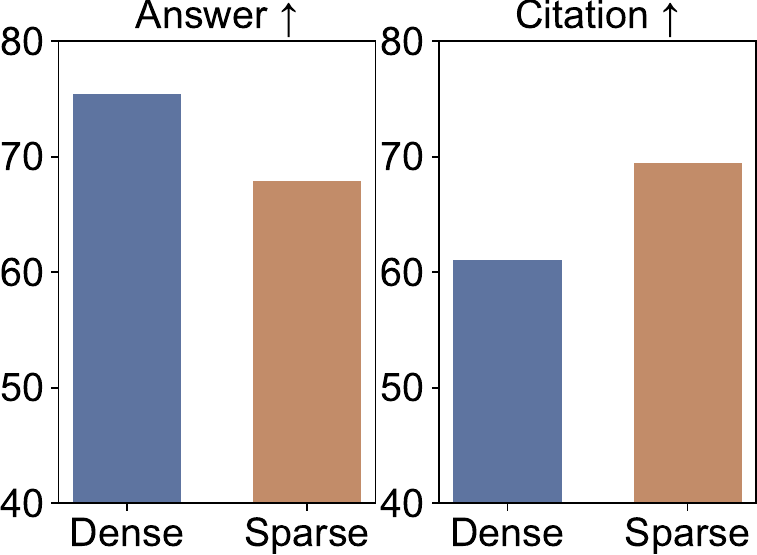}
        \caption{Multi hop WebQA}
        \label{fig:multi-webqa}
    \end{subfigure}
    \caption{Dense RAG vs Sparse RAG in terms of answer accuracy and attribution capability. \textbf{(a)},\textbf{(b)}: Sparse RAG shows superiority on both metrics; \textbf{(c)},\textbf{(d)}: Sparse RAG still excels on attribution performance, but has worse answer quality.}
    \label{fig:sparserag}
\end{figure}

The second experiment compares a state-of-the-art sparse RAG~\cite{zhu2025accelerating} with dense RAG (w/o context filtering) on simultaneously answering the question and citing the multi-modal contexts.  
Figure~\ref{fig:single-mmqa} and~\ref{fig:single-webqa} show that the sparse RAG produces better answers and citations than dense RAG on single-hop questions. However, in Figure~\ref{fig:multi-mmqa} and~\ref{fig:multi-webqa}, directly applying sparse RAG for multi-hop questions leads to lower answer quality. It implies that sparse RAG drops golden contexts and suffers from hallucinations. 

{\noindent \textbf{Discussion}}. 
We emphasized that sparse RAGs generate better quotes than dense RAGs. 
We found that the existing sparse RAGs (a.k.a. reranking models) excel in single-hop question answering and citation generation, but face challenges for multi-hop questions. 
Thus, we propose a novel RAGQ, namely QDRAG, which is superior in both single-hop and multi-hop scenarios.

\subsection{QDRAG Overview}
As shown in Figure~\ref{fig:intro}, QDRAG comprises three main modules: query decomposer, context reranker, and answer generator. The decomposer derives multiple subquestions, and the reranker purifies the retrieved data by their relevance to subquestions. 
Afterwards, the answers and citations are generated auto-regressively by the MLLM.

\subsection{QD-C: Context-Guided Question Decomposition}

QD-C defines two operators for query decomposition: entity replacement and open decomposition for inter-dependent and independent subquestions, respectively.

\subsubsection{Entity Replacement.} 
We instruct the MLLM to identify any referential expressions (e.g., ``The voice artist of the girl mammoth in \emph{Ice Age: Continental Draft}'') within the question. Then, the MLLM utilizes the external knowledge to substitute references with entities (e.g., ``Keke Palmer''). 
The operator produces two subquestions for each replacement. One is ``Is $\langle$reference$\rangle$ equivalent to $\langle$entity$\rangle$?'', the other is the new question after substitution.

\subsubsection{Open Decomposition.} The MLLM is tasked with allocating queried entities to multiple subquestions to enable answering with a single context. For instance, for question ``Which one released first, A or B?'', the operator produces two subquestions regarding the released year of two entities. This approach safeguards against the reranker disregarding contexts that contain partial information (e.g., the release year) without explicit comparison. 

QD-C eliminates manually-written in-context examples by the well-defined operators. Moreover, the two operators can be implemented into a single MLLM inference for each multi-hop question, which makes QD-C cost-effective and plug-and-play.

\subsection{QD-R: Efficient and Generalizable QD Alignment} \label{sec:qdr}

\subsubsection{Motivation.}
Similar to other single-round retrieval RAGs, QD-C assumes that the retrieved documents contain sufficient information to decompose the question. However, a single round of lightweight retrieval often fails to recall the complete set of golden contexts~\cite{yu2024rankrag}, which leads to cascading errors that degrade answer and citation quality.

\subsubsection{QD-R.}

We propose retrieval-guided QD (QD-R), a new query decomposition method with multi-round retrievals. 
Firstly, QD-R takes only the query as input and performs zero-shot query decomposition. The output of QD-R is simply a decomposition plan comprising multiple subquestions. QD-R allows the special token ``$\langle$Ans\_of\_Qi$\rangle$'' in subquestions, which means that the content of the subquestion depends on the answer of previous subquestions. Then, QD-R resolves the placeholders via RAG and replaces the special tokens with the answers to that question. Finally, different from QD-C, each resolved subquestion retrieves its own set of contexts for subsequent reranking.

\begin{figure}[t]
    \centering
    \includegraphics[width=0.99\linewidth]{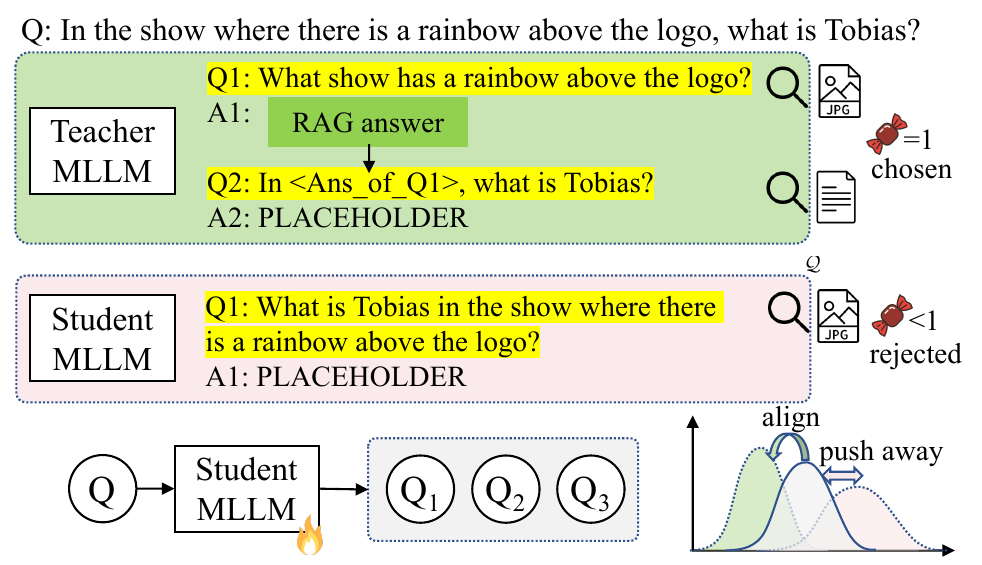}
    \caption{QD-R samples multiple MLLM outputs and uses the recall of golden contexts as reward to train the model.}
    \label{fig:qdr}
    \vspace*{-2mm}
\end{figure}

\subsubsection{QDARF.}

QD-R regularizes its decompositions by \textbf{A}ligning MLLM from \textbf{R}etrieval \textbf{F}eedback (QDARF). As shown in Figure~\ref{fig:qdr}, we sample multiple decomposition plans from a teacher MLLM (Qwen2.5VL-32B) and a student MLLM (Qwen2.5VL-7B). The reward of a decomposition plan is defined as the recall of golden contexts by all its subquestions. We adopt the direct preference optimization (DPO) loss function to finetune the MLLM:

\begin{equation}
\begin{aligned}
\mathcal{L}=&-E_{(x,y_w,y_l)\sim\mathcal{D}}\left[\log\sigma\left(\beta\log\frac{\pi_\theta\left(y_w\mid x\right)}{\pi_{\mathrm{ref}}\left(y_w\mid x\right)}\right.\right.\\&-\beta\log\frac{\pi_\theta\left(y_l\mid x\right)}{\pi_{\mathrm{ref}}\left(y_l\mid x\right)}\biggr)\biggr],
\end{aligned}   
\label{eq:dpo} 
\end{equation}
where $y_w$ are $y_l$ are the positive (winning) and the negative (losing) decomposition plans, respectively. The decompositions with the \emph{highest} reward values are deemed as positive, while those with the \emph{lowest} reward values are treated as negative. $\pi_{\text{ref}}$ is the initial MLLM and $\pi_\theta$ is the MLLM to optimize. 
Optimizing Eq.~\eqref{eq:dpo} increases the margin between the positive and negative plans. Consequently, the subquestions would have a larger recall rate of golden contexts after finetuning, which benefits the downstream reranking task.

\subsection{Context Reranking}

QDRAG evaluates the context's relevance to each subquestion and the raw question. The relevance score of a context $C$ to a subquestion $Q_i$ is defined as the probability of ``Yes'' as the next token under the reranking prompt $R$:
\[
    \mathcal{S}(C|Q_i)=\mathcal{P}(\text{``Yes''}|Q_i,C,R)
\]
Each subquestion retains a certain number of top-relevant contexts. 

\subsection{Adaptive Budget Allocator}

Given a budget $N$, how to distribute the number of reserved contexts for each subquestion? We propose an entropy-based method. The idea is that higher entropy implies that the MLLM is less certain about selecting the best supporting document, hence requiring more contexts to be retained for that subquestion.
Formally, 
\[
\begin{aligned}
\mathcal{P}(C|Q_i)&=\frac{\exp(\mathcal{S}(C|Q_i))}{\sum_{C'\in \mathcal{C}}\exp(\mathcal{S}(C'|Q_i))} \\
\mathcal{E}(Q_i)&=-\sum_{C'\in \mathcal{C}}\mathcal{P}(C'|Q_i)\log \mathcal{P}(C'|Q_i)
\end{aligned}
\]

Then, the number of reserved contexts allocated to $Q_i$ is 
\[
n(Q_i)=N\cdot\frac{\mathcal{E}(Q_i)}{\sum_{Q_j\in\mathcal{Q}}\mathcal{E}(Q_j)}
\]


\section{Experiment} \label{sec:exp}

\definecolor{color1}{HTML}{edf6ee}
\definecolor{color2}{HTML}{edf3fd}
\definecolor{color3}{HTML}{FEFAE0}
\newcolumntype{a}{>{\columncolor{color1}}r}
\newcolumntype{f}{>{\columncolor{color2}}r}
\newcolumntype{d}{>{\columncolor{color3}}r}
\newcolumntype{e}{>{\columncolor{gray!10}}r}

\begin{table*}[t]
    \small
    \centering
    \setlength{\tabcolsep}{7.6pt}

\begin{tabular}{@{}llrrrrrrrrrrrr@{}}
\toprule
\multirow{3}{*}{Model} & \multirow{3}{*}{Model Size} & \multicolumn{6}{c}{In-domain} & \multicolumn{6}{c}{Out-of-domain} \\
 &  & \multicolumn{3}{c}{MMQA}  & \multicolumn{3}{c}{WebQA} & \multicolumn{3}{c}{InfoSeek} & \multicolumn{3}{c}{E-VQA}\\ \cmidrule(l){3-14} 
 &  & EM & LLM & Cite & EM & LLM & Cite & EM & LLM & Cite & EM & LLM & Cite \\ 
\midrule
\multicolumn{14}{c}{\textit{Retrieval-Generation}} \\
\midrule
Gemini-Pro & - & 43.2 & 34.2 & \textbf{68.4} & 55.1 & 26.6 & \underline{64.6} & 9.8 & 0.0 & 0.9 & 6.1 & 3.4 & 23.6 \\
Gemini-Pro$_{\text{CLIP}}$ & - & - & - & 60.4 & - & - & 57.9 & - & - & 0.4 & - & - & 27.9 \\
\hdashline
Qwen2.5VL & 7B & 42.4 & 40.6 & 38.2 & 57.8 & 41.5 & 33.0 & 42.5 & \underline{36.7} & 59.6 & 10.4 & 7.8 & 18.5 \\
Qwen2.5VL$_{\text{CLIP}}$ & 7B & - & - & 47.2 & - & - & 35.5 & - & - & 58.4 & - & - & 26.5 \\
InternVL3 & 14B & 49.1 & \underline{50.1} & 50.9 & 58.3 & 43.6 & 44.5 & 36.0 & 30.5 & 52.2 & 8.5 & 7.3 & 15.9 \\
InternVL3$_{\text{CLIP}}$ & 14B & - & - & 47.6 & - & - & 39.7 & - & - & 52.2 & - & - & 16.6 \\
\midrule
\multicolumn{14}{c}{\textit{Retrieval-Reranking-Generation}} \\
\midrule
CLIP & 428M/7B & 35.0 & 32.4 & 39.5 & 50.9 & 22.9 & 36.8 & 40.7 & 32.4 & 56.0 & 9.2 & 6.5 & 16.1 \\
MMEmbed & 7B & 43.6 & 41.2 & 51.4 & 55.9 & 36.7 & 44.7 & 41.8 & 35.6 & 60.9 & 9.5 & 6.9 & 22.0 \\
Sparse-RAG & 7B & 45.9 & 46.1 & 48.1 & 57.6 & 46.1 & 39.0 & \underline{44.5} & 35.1 & 58.9 & 14.6 & 11.6 & 15.7 \\
RagVL & 13B/7B & 42.3 & 40.8 & 50.9 & 56.1 & 38.9 & 43.4 & - & - & - & - & - & - \\
\midrule
\multicolumn{14}{c}{\textit{Retrieval-QD-Reranking-Generation}} \\
\midrule
LlamaIndex & 7B & 43.3 & 42.1 & 50.0 & 56.4 & 40.8 & 44.1 & 10.4 & 2.1 & 3.2 & 13.1 & 10.7 & 20.0 \\
QDRAG-C (ours) & 7B & 43.3 & 42.2 & 52.6 & 56.7 & 40.2 & 45.8 & 42.2 & 35.4 & 62.1 & 9.6 & 7.5 & 21.2 \\
QDRAG-C (ours)& 14B & 46.8 & 46.9 & 55.8 & 57.4 & 39.7 & 48.6 & 36.9 & 28.7 & 56.5 & 6.5 & 5.5 & 15.3 \\
\midrule
\multicolumn{14}{c}{\textit{QD-Retrieval-Reranking-Generation}} \\
\midrule
Self-Ask & 7B & 42.9 & 41.9 & 49.9 & 56.2 & 40.1 & 43.8 & 42.7 & 34.5 & 60.6 & 9.7 & 8.4 & 21.1 \\
SearChain & 7B & 43.7 & 42.6 & 49.3 & 56.3 & 40.6 & 45.1 & 42.4 & 34.3 & 64.0 & 10.0 & 8.0 & 21.7 \\
\hdashline
QDRAG-R (ours)& 7B & \underline{49.2} & 46.3 & 57.5 & \underline{61.7} & 48.0 & 59.9 & \textbf{46.0} & \textbf{38.3} & \textbf{72.9} & \textbf{16.2} & \textbf{12.8} & \underline{33.8} \\
w/o DPO & 7B & 46.7 & 45.4 & 53.8 & 61.0 & \underline{48.3} & 56.8 & 42.0 & 35.2 & 66.3 & 13.4 & 10.6 & 28.9 \\
QDRAG-R (ours)& 14B & \textbf{53.1} & \textbf{52.8} & \underline{65.6} & \textbf{62.6} & \textbf{53.5} & \textbf{70.9} & 39.8 & 32.4 & \underline{68.7} & \underline{15.7} & \underline{12.3} & \textbf{40.4} \\ 
\bottomrule
\end{tabular}
\caption{Overall in-domain and out-of-domain test, where best and second-best results are highlighted in \textbf{bold} and \underline{underlined}, respectively. The retriever for all models is CLIP-ViT-L/14@336px.  Single model size means that QD/Reranking/QA use the same MLLM, while multiple model sizes (e.g., 13B/7B) means the reranker model size and the QD/QA model size, respectively. }
    \label{tab:main}
    \vspace*{-1mm}
\end{table*}

\begin{table}[t]
    \centering
\begin{tabular}{@{}lrrrrrr@{}}
\toprule
& \multicolumn{3}{c}{MMQA} & \multicolumn{3}{c}{WebQA} \\
& EM & LLM & Cite & EM & LLM & Cite \\ \midrule
RAG & 67.7 & 64.1& 51.3 & \textbf{70.7} &\textbf{64.0}& 60.9 \\
MMEmbed & 69.7& 65.1& 71.5 & 68.0 &57.1& 69.4\\
SparseRAG & 67.9& 66.1& 71.9 & 68.5 & 60.9&65.3 \\
LlamaIndex & 68.0&65.7 & 69.9 & 69.6 &61.2& 72.1\\
QDRAG-C & \textbf{70.7} &\textbf{66.4}& \textbf{74.7} & \textbf{70.7} &63.7& \textbf{75.1} \\
\bottomrule
\end{tabular}
\caption{Comparison of single-round RAGs with 7B retriever}
    \label{tab:7bret}
    \vspace*{-3mm}
\end{table}

\subsection{Experimental Setup}

\subsubsection{Datasets.} We conduct evaluations on four multi-modal question answering datasets: MMQA~\cite{talmor2021multimodalqa}, WebQA~\cite{Chang2021WebQAMA}, InfoSeek~\cite{chen-etal-2023-pre-trained} and E-VQA~\cite{mensink2023encyclopedic}. 
MMQA and WebQA consist of 1,196 and 4,967 testing queries in text form, respectively. Each query necessitates one or more image or text fragments for a complete answer. InfoSeek and E-VQA questions task the models with entity recognition in images and retrieval of relevant textual information for answering.

\subsubsection{Evaluation Metrics.} We evaluate the RAG outputs from two dimensions: answer quality and context reference faithfulness. For answer quality, we use exact match (EM) and scores graded by LLM-judge. EM measures the ratio of keywords in the ground-truth answer that are exactly mentioned in the model's response. LLM-judge prompts an MLLM $\psi$ to grade the correctness and comprehensiveness of RAG outputs given the golden answers. 
For context reference capability, following existing work~\cite{song2025measuring}, we perform AutoAIS evaluation using an off-the-shelf MLLM $\psi$. The inputs to $\psi$ include the user question, the model-generated response, and the cited contexts. The AutoAIS output is a True/False label indicating whether the aggregated contexts support the model response. In this paper, we set $\psi$ as Qwen2.5VL-32B~\cite{qwen2.5-VL}.

\subsubsection{Baselines.} 
As shown in Table~\ref{tab:main}, the main experiment contains four categories of baselines. The first category comprises dense RAGs with recent MLLM backbones, including the commercial 
Google Gemini2.5-Pro~\cite{gemini} and the open-sourced Qwen2.5-VL~\cite{qwen2.5-VL} and InternVL-3~\cite{zhu2025internvl3}.
Following \cite{huang-etal-2024-learning}, this category also contains post-hoc retrieval baselines that select reference contexts with the CLIP retriever. The second category uses multi-modal rerankers to implement sparse RAGs, including CLIP~\cite{pmlr-v139-radford21a}, RagVL~\cite{chen2024mllmstrongrerankeradvancing}, MMEmbed~\cite{lin2025mmembed}, and SparseRAG~\cite{zhu2025accelerating}. The third category is QD models with single-round retrieval, including LlamaIndex\footnote{\url{https://github.com/run-llama/llama_index}} and the proposed QDRAG-C. The last category compares the query decomposition methods under the QDRAG-R pipeline. The baselines include Self-Ask~\cite{press-etal-2023-measuring} and SearChain~\cite{xu2024search}. In ablation study, we further compare QDRAG with rewriting models~\cite{mao-etal-2024-rafe} and the instruction-finetuned  RAGQs~\cite{huang-etal-2024-learning}. 

\vspace*{-2mm}
\subsubsection{Implementation Details.}
If not specified, the backbone of 7B RAGs and 14B RAGs are Qwen2.5-VL-7B~\cite{qwen2.5-VL} and  InternVL3-14B~\cite{zhu2025internvl3}, respectively. The reranking models reserve top-$5$ contexts. 
We adopt low-rank adaptation (LoRA)~\cite{hu2022lora} when finetuning the MLLM. 
The model inference is implemented using the vLLM v0.9.2 framework~\cite{kwon2023efficient}. The experiments are conducted on a Ubuntu server with $2\times\text{A100 80B}$ NVIDIA GPUs. For detailed hyperparameters and prompts, we refer the reader to the appendix.

\begin{table}[t]
    
    \centering
\begin{tabular}{@{}lrrrrrr@{}}
\toprule
& \multicolumn{3}{c}{MMQA} & \multicolumn{3}{c}{WebQA} \\
& EM & LLM & Cite & EM & LLM & Cite \\ \midrule
RaFe-C & 43.3 &42.2& 49.9 & 55.6 &40.3& 44.4 \\
QD-C & 43.3 &42.2& 52.6 & 56.7 &40.2& 45.8 \\ \midrule
RaFe-R & 43.6  &42.0& 49.3 & 55.7 &38.7& 42.3 \\
RaFe-R$_{\text{DPO}}$ & 43.8 &42.7& 50.2 & 57.4 &41.7& 48.4 \\
QD-R & 46.7 &45.4& 53.8 & 61.0 &\textbf{48.3}& 56.8 \\
QD-R$_{\text{DPO}}$ & \textbf{49.2} &\textbf{46.3}& \textbf{57.5} & \textbf{61.7} &48.0& \textbf{59.9} \\
\bottomrule
\end{tabular}
\caption{Compare QD with rewriting models}
    \label{tab:rewrite}
\end{table}

\begin{figure}[t]
    \centering
    \begin{subfigure}{0.45\linewidth}
        \centering
        \includegraphics[width=\linewidth]{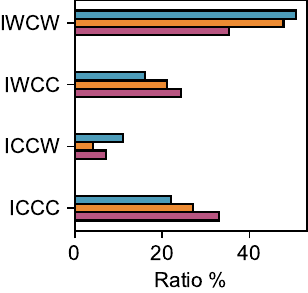}
        \caption{MMQA}
    \end{subfigure}
    \quad
    \begin{subfigure}{0.45\linewidth}
        \centering
        \includegraphics[width=\linewidth]{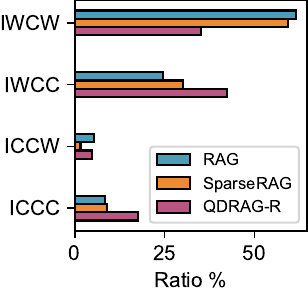}
        \caption{WebQA}
    \end{subfigure}
    \caption{Failure case analysis. \textbf{IWCW} $\downarrow$: Input wrong context and cite incorrectly; \textbf{IWCC} $\uparrow$: Input wrong context but cite correctly; \textbf{ICCW} $\downarrow$: Input correct context but cite incorrectly; \textbf{ICCC} $\uparrow$: Input correct context and cite correctly.}
    \label{fig:failure}
    \vspace*{-2mm}
\end{figure}

\subsection{Main Results}

{\noindent \textbf{Context Filtering Boosts Citation Quality.}} As shown in Table~\ref{tab:main}, 
reranking RAGs consistency show improvements over both prompt-based and post-hoc retrieval baselines. This highlights our proposition that the sparse context helps MLLMs better attribute their answer to supporting contexts. Comparing CLIP with other rerankers reveals that relying solely on the retrieval score for reranking is suboptimal.

{\noindent \textbf{QDRAG-R Achieves Significant Improvements on In-domain and Out-of-Domain Tests.}} Our QDRAG-R models are trained on the training splits of MMQA and WebQA. InfoSeek and E-VQA have image inputs in all user queries, while MMQA and WebQA only contain pure-text queries. The RAG knowledge bases are also different. Therefore, our training data exhibits out-of-domain characteristics. Nonetheless, QDRAG-R demonstrates a substantial performance advantage on both answer quality and attribution capability over existing reranking RAGs and QD-reranking RAGs. The QDRAG-R-14B outperforms the proprietary MLLM Gemini2.5-pro in most cases. 

{\noindent \textbf{QDRAG-C Excels with More Powerful Retrievers.}} In Table~\ref{tab:main}, QDRAG-C shines at resource attribution but is less splashy for answer quality. This highlights our proposition that QDRAG-C is potentially sensitive to initial retrieval quality. In Table~\ref{tab:7bret}, we replace the lightweight CLIP retriever with MMEmbed-7B. Then QDRAG-C demonstrates improvements in both answer quality and attribution capability. 

\subsection{Analysis}

\subsubsection{Rewriting-R vs QD-R}

Rewriting is another important query processing strategy. 
We compare QD-reranking with the recent query rewriting method RaFe~\cite{mao-etal-2024-rafe} by adapting it for RaFe-reranking. As shown in Table~\ref{tab:rewrite}, in single-round retrieval setup, QD-C shows improvement compared to RaFe-C. In multi-round retrieval setup, we construct the RaFe training dataset similar to that of QD-R. The only difference is that RaFe instructs the MLLM to rewrite the query. The results show that query decomposition is indeed an essential step for RAG trustworthiness and answer quality. 

\begin{table}[t]
    
    \centering
\begin{tabular}{@{}llrrrr@{}}
\toprule
\multirow{2}{*}{Pipe} & \multirow{2}{*}{FT task} & \multicolumn{2}{c}{InfoSeek} & \multicolumn{2}{c}{EVQA} \\
& & LLM & Cite & LLM & Cite \\ \midrule
\multirow{3}{*}{RAG} & - & 36.7 & 59.6& 7.8& 18.5\\
 & RAGQ & 32.9 & 55.6& 6.4& 18.7\\
 & QD-R & 36.6& 62.3& 7.8& 19.1\\ \midrule
\multirow{3}{*}{QDRAG-R} & - & 35.2& 66.3& 10.6& 28.9\\
 & RAGQ & 33.0& 68.4& 8.6& 26.0\\
 & QD-R & \textbf{38.3}& \textbf{72.9}& \textbf{12.8}& \textbf{33.8}\\ 
\bottomrule
\end{tabular}
\caption{Out-of-domain test: Front vs QDRAG}
    \label{tab:dpo_task}
\end{table}

\begin{table}[t]
    \centering
    \setlength{\tabcolsep}{7.6pt}
\begin{tabular}{@{}cccc@{}}
\toprule
FT task & Feedback & Auto & Cost \\ \midrule
RAGQ & LLM-judge & $\checkmark$ & 28h\\ 
QD-R & Retrieval & $\checkmark$ & 2h\\
\bottomrule
\end{tabular}
\caption{Cost of feedback when building training data}
    \label{tab:feedback_cost}
    \vspace*{-2mm}
\end{table}

\subsubsection{Finetuning Task}

We compare QDRAG with a representative RAGQ finetuning method: Front~\cite{huang-etal-2024-learning}. Front retrieves contexts for training questions and uses a teacher model and a student model to generate positive and negative answers with citations, respectively. For fair comparison, we use the same teacher model and student model as QDRAG. The results are shown in Table~\ref{tab:dpo_task}. Similar to published results~\cite{huang-etal-2024-learning,song2025measuring}, we observe descending answer accuracy of RAGQ on out-of-domain datasets. QDRAG shows a significantly better generalization capability on out-of-domain datasets.

In Table~\ref{tab:feedback_cost}, we compare the annotation cost of Front and QDRAG. The results show that the retrieval feedback is significantly more efficient than MLLM feedback. 

\subsubsection{What Makes QD-R Better?}

In Figure~\ref{fig:failure}, we divide the inputs into four cases for each model. Our key findings are: (1) most incorrect references are caused by incorrect context inputs, which highlights the importance of QD-based context reranking; (2) QD-R significantly reduces the failure cases.

\subsubsection{Ablation Study}

\begin{table}[t]
    
    \centering
\begin{tabular}{@{}cccrrrrrr@{}}
\toprule
\multirow{2}{*}{D} & \multirow{2}{*}{R} & \multirow{2}{*}{S} & \multicolumn{3}{c}{MMQA}& \multicolumn{3}{c}{WebQA}  \\
& & & EM & LLM & Cite & EM & LLM & Cite \\ \midrule
$\checkmark$ & $\checkmark$ & $\checkmark$ & 43.3 &42.2& 52.6 & 56.7 &40.2& 45.8\\ \midrule
$\checkmark$ &  & $\checkmark$ &43.3&41.8& 51.0 & 56.8 &40.8& 45.4\\
 & $\checkmark$ & $\checkmark$ &43.8&41.4& 50.9 & 56.9 &39.4& 45.2\\
 $\checkmark$ & $\checkmark$ &  &44.0&42.0&52.0 & 56.5 &40.7& 45.8\\
 &  &  & 43.5&42.0& 48.8 & 55.8 &39.8& 41.9\\
\bottomrule
\end{tabular}
\caption{Ablation on QDRAG-C}
    \label{tab:ablation_qdc}
\end{table}

\begin{table}[!htbp]
    
    \centering
\begin{tabular}{@{}cccrrrrrr@{}}
\toprule
\multirow{2}{*}{M} & \multirow{2}{*}{W} & \multirow{2}{*}{S} & \multicolumn{3}{c}{MMQA}& \multicolumn{3}{c}{WebQA}  \\
& & & EM & LLM & Cite & EM & LLM & Cite \\ \midrule
$\checkmark$ & $\checkmark$ & $\checkmark$ & 49.2 &46.3& 57.5 & 61.7 &48.0& 59.9\\ \midrule
$\checkmark$ &  & $\checkmark$ & 46.9 &44.6& 56.9 & 61.1 &48.1& 56.9\\
 & $\checkmark$ & $\checkmark$ & 46.4 &45.4& 54.3 & 61.6 &48.3& 58.3\\
 $\checkmark$ & $\checkmark$ &  & 48.8 &46.0& 56.5 & 61.9 &47.8& 60.6\\
 &  &  & 46.9 &45.0& 54.8 & 61.0 &48.1& 56.9\\
\bottomrule
\end{tabular}
\caption{Ablation on QDRAG-R Traing Data}
    \label{tab:ablation_qdr}
\end{table}

We conduct a set of ablations of QDRAG to identify which factors play key roles. We examine the benefits of the replacement (R) and the decomposition (D) operators in QD-C. From Table~\ref{tab:ablation_qdc}, we observe that both of them are critical for effective query decomposition. Compared with no query decomposition, QD-C shows a large performance gain. We test the contributions of QD-R training datasets in Table~\ref{tab:ablation_qdr}. The result shows QD-R's scalability w.r.t. the training data size. In both models, the adaptive context number assignment module (S) provides performance gains.

\begin{figure}[!htbp]
    \centering
    \includegraphics[width=\linewidth]{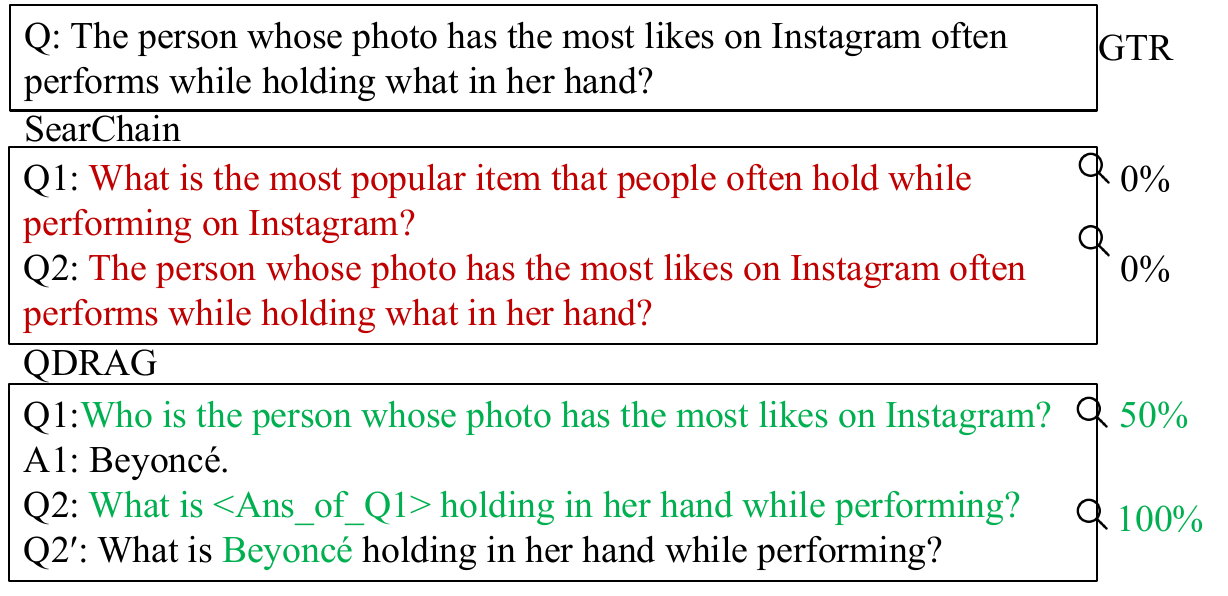}
    \caption{Case study: QDRAG generates better query decompositions. GTR: Golden context Recall.}
    \label{fig:decomp}
    \vspace*{-2mm}
\end{figure}

\subsubsection{Efficiency \& Cost}

\begin{figure}[t]
    \centering
    \begin{subfigure}{0.99\linewidth}
        \centering
        \includegraphics[width=\linewidth]{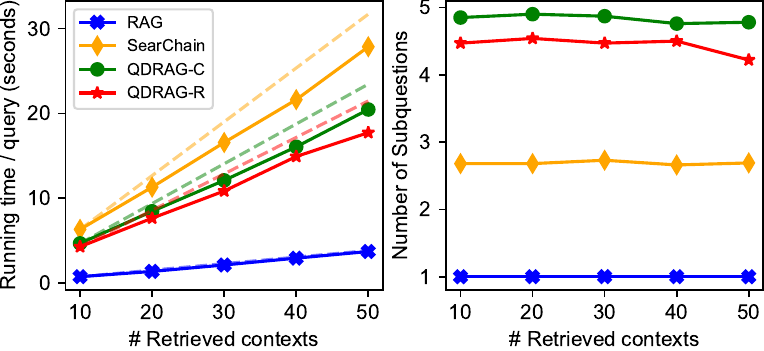}
        \caption{Analysis of efficiency and scalability}
        \label{fig:efficient}
    \end{subfigure} \\
    \begin{subfigure}{0.99\linewidth}
        \centering
            \begin{tabular}{@{}llrrrr@{}}
            \toprule
            Method& $\#n$ & $\#m$ & $\#r$ & Cost $\downarrow$ \\ \midrule
            SearChain &25715& 391& 10& 0.070\\
            QDRAG-C &4810& 451& 1& 0.019\\ 
            QDRAG-R &12660& 237& 4&0.035\\ 
            \bottomrule
            \end{tabular}
        \caption{Analysis of QD costs per query in dollars}
        \label{fig:cost}
    \end{subfigure}
    \caption{Efficient \& cost analysis}
    \label{fig:efficiency_and_cost}
\end{figure}

In Figure~\ref{fig:efficient}, we analyze the efficiency and scalability w.r.t. the number of retrieved contexts at each retrieval round. The dashed lines indicate the linear growth of running time. 
Compared with baselines, our method substantially boosts task performance without significantly increasing time consumption compared to the baselines. Notably, QDRAGs generate more effective subquestions with less running time.

In Figure~\ref{fig:cost}, we analyze the costs of query decomposition in dollars. The reported metrics include the average number of input tokens ($n$), output tokens ($m$), and MLLM calls ($r$). The overall cost is calculated by $\sum_r(n_rp_1+m_rp_2)$, where $p_1$ and $p_2$ denote the unit price of input and output tokens of Gemini2.5-pro, respectively. QDRAG-C only incurs a single MLLM call to make it cost-effective. QDRAG-R has a slightly higher cost but has higher performance.

\begin{figure}[t]
    \centering
    \begin{subfigure}{0.465\linewidth}
        \centering
        \includegraphics[width=\linewidth]{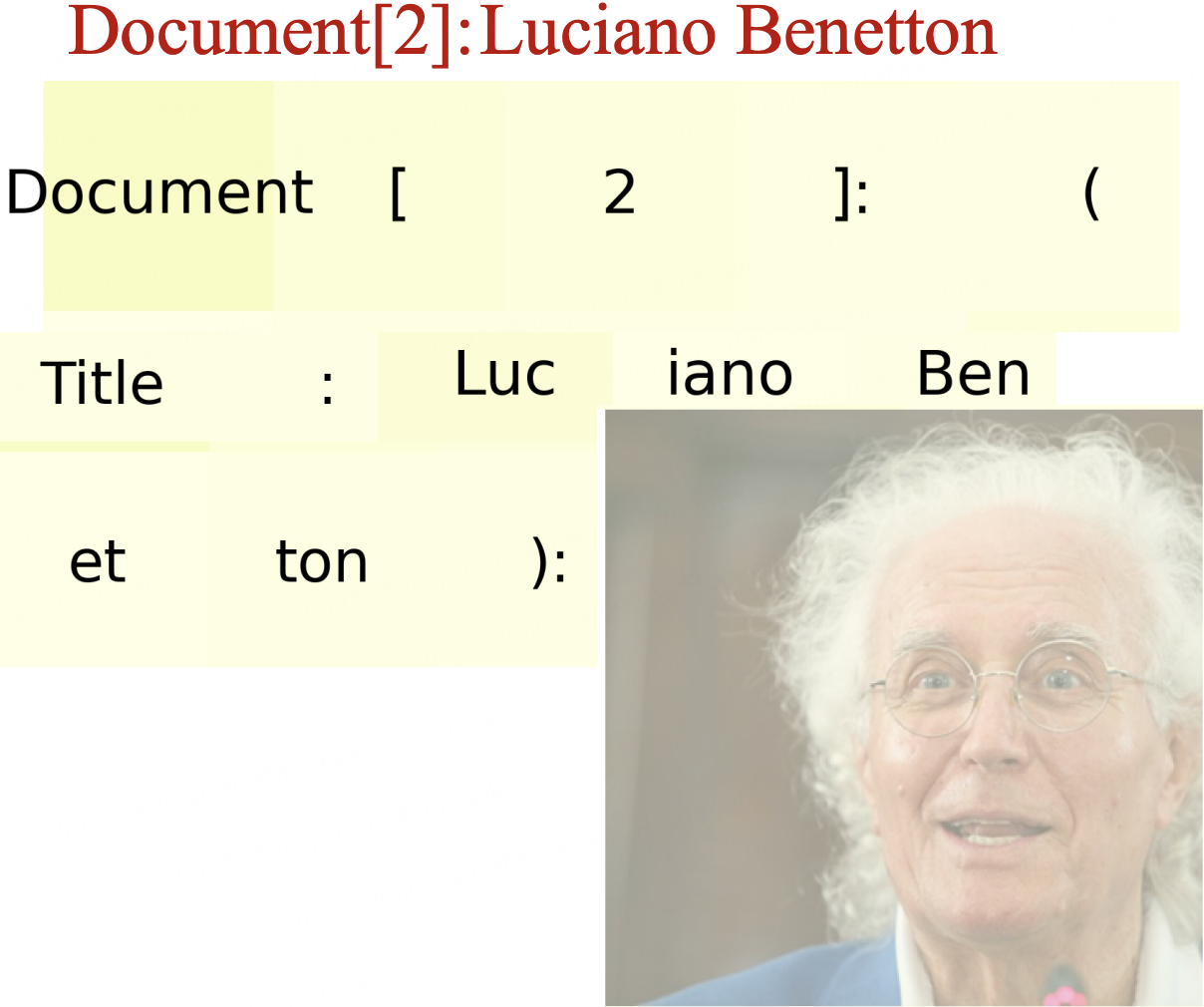}
        \caption{Dense context RAG}
    \end{subfigure}
    \quad
    \begin{subfigure}{0.465\linewidth}
        \centering
        \includegraphics[width=\linewidth]{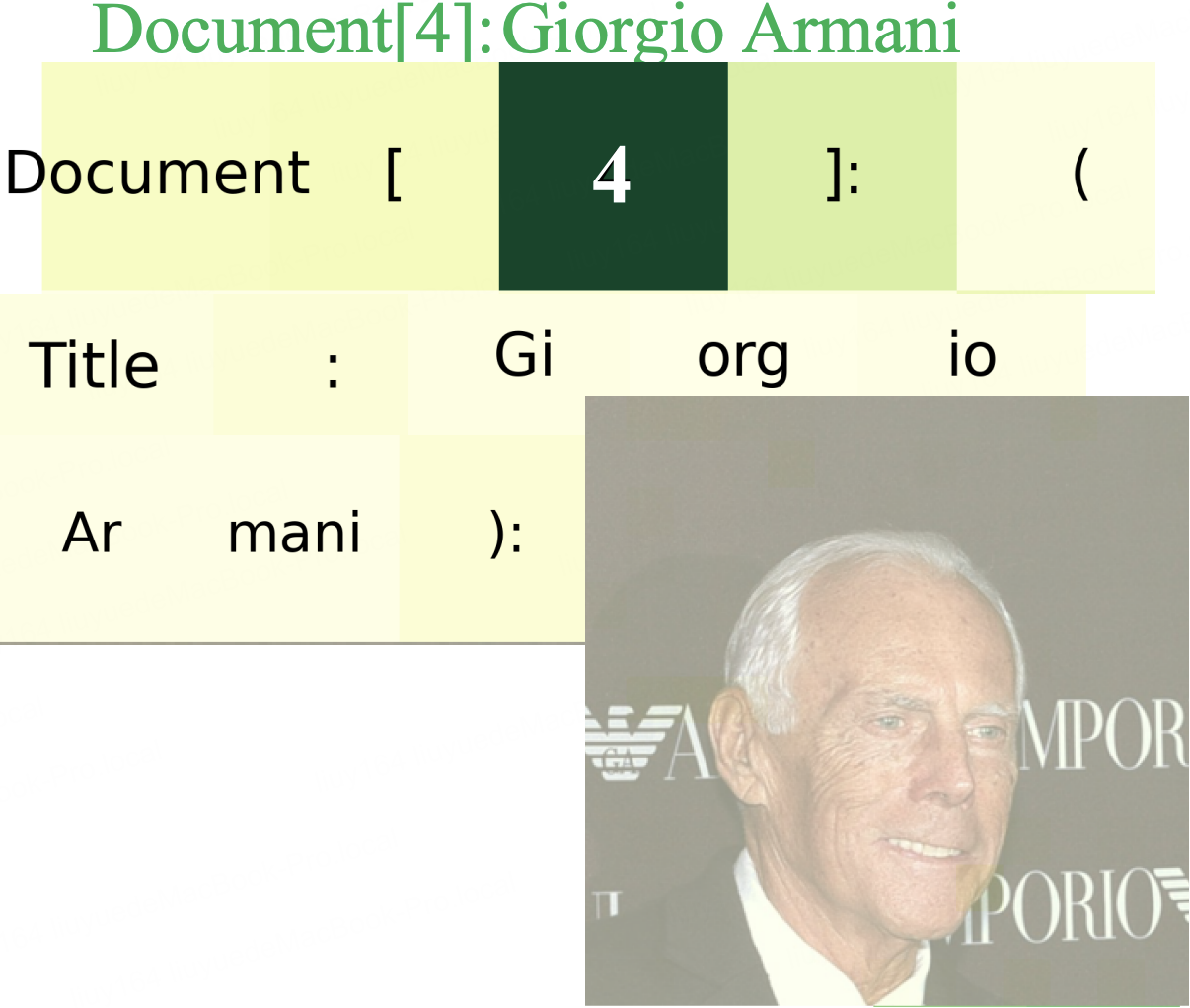}
        \caption{QDRAG}
    \end{subfigure}
    \caption{Visualization of image contexts with the largest attention values when MLLM generates the references after ``What color is Giorgio Armani's hair? White. According to Document \textvisiblespace''}
    \label{fig:attn}
    \vspace*{-2mm}
\end{figure}

\subsubsection{Case Study}
In Figure~\ref{fig:decomp}, the baseline decompositions have almost the same level of difficulty with the input question. Our QDRAG successfully simplifies the question step by step.
The baselines prompt an MLLM to decompose the query in the open exploration space. This may lead to unsolvable sub-queries or rewritings rather than really reducing the question complexity. In our method, we enable MLLM-IR interaction by preference optimization. The MLLM in QDRAG tends to generate reasonable subquestions that have large rewards, i.e., the recall rate of golden contexts. This benefits downstream tasks of question answering and contextual referencing.

In Figure~\ref{fig:attn}, we visualize the image context attention that has the largest attention values on the context ID (e.g., Document [4]) when the QA MLLM generates references. The dense RAG spends large attention on a photo of an incorrect person who also has white hair, while QDRAG is more confident and wise when generating reference IDs. 

\section{Conclusion}

In this work, we propose QDRAG, a novel RAG framework to enhance attributed RAG. We present that the sparse context helps LLM concentrate on important contexts and point out the limitation of existing sparse RAGs. We propose two new query-decomposition methods. In particular, QD-C is guided by initial contexts, while QD-R is guided by retrieval rewards. QDRAG successfully addresses the three challenges of existing RAGQs. Extensive experiments are conducted to validate the generalizability and scalability of QDRAG.

\section{Acknowledgments}
The research work described in this paper was supported by Hong Kong Research Grants Council (grant\# T43-513/23-N, T22-607/24N). It was partially conducted in JC STEM Lab of Data Science Foundations funded by The Hong Kong Jockey Club Charities Trust. Shimin Di’s work is supported by National Science Foundation of China (NSFC) under Grant No. 62506075.

\bibliography{references}

\end{document}